\documentclass[11pt,a4paper]{article}

\usepackage[affil-it]{authblk}
\usepackage[margin=3.0cm]{geometry}
\usepackage{amsmath,amssymb,amsfonts,upref}
\usepackage{amsthm}
\usepackage{mathtools}
\usepackage{tabularx}

\title{Automatic Programming of Cellular Automata and Artificial Neural Networks Guided by Philosophy}
\author[1]{Patrik Christen\footnote{Corresponding author: patrik.christen@fhnw.ch. The authors contributed equally to this work.}}
\affil[1]{Institute for Information Systems, FHNW University of Applied Sciences and Arts Northwestern Switzerland, Riggenbachstrasse 16, 4600 Olten, Switzerland}
\author[2]{Olivier Del Fabbro}
\affil[2]{Chair for Philosophy, ETH Zurich, Clausiusstrasse 49, 8092 Zurich, Switzerland}
\date{{\small Appeared in \textit{Rolf Dornberger, editor, New Trends in Business Information Systems and Technology: Digital Innovation and Digital Business Transformation, pages 131--146. Springer, Cham, 2020.}}\\ \vspace{22pt}10 May 2019 (last revised 31 August 2020)}

\begin{document}

\maketitle

\abstract{Many computer models have been developed and successfully applied. However, in some cases, these models might be restrictive on the possible solutions or their solutions might be difficult to interpret. To overcome this problem, we outline a new approach, the so-called allagmatic method, that automatically programs and executes models with as little limitations as possible while maintaining human interpretability. Earlier we described a metamodel and its building blocks according to the philosophical concepts of structure and operation. They are entity, milieu, and update function that together abstractly describe a computer model. By automatically combining these building blocks in an evolutionary computation, interpretability might be increased by the relationship to the metamodel, and models might be translated into more interpretable models via the metamodel. We propose generic and object-oriented programming to implement the entities and their milieus as dynamic and generic arrays and the update function as a method. We show two experiments where a simple cellular automaton and an artificial neural network are automatically programmed, compiled, and executed. A target state is successfully reached in both cases. We conclude that the allagmatic method can create and execute cellular automaton and artificial neural network models in an automated manner with the guidance of philosophy.}



\newpage
\section{Introduction}

Computer modelling has become an important approach in many scientific disciplines that study their respective systems. It allows to experiment digitally with the modelled system and as such to explore and test possible explanations for its functioning. This is especially important and useful for the understanding of complex systems, where the system's behaviour emerges from the interaction of local elements. Most of the computer models that have been proposed so far have their specific applications or type of problems for which they are most appropriate. E.g. cellular automata are particularly suitable for simulating complex behaviour in general \cite{Wolfram1984} and development and growth processes in particular as they occur in materials \cite{DingGuo2002} and biology \cite{JiaoTorquato2011}. Due to their capability to model complex behaviour in general, cellular automata have been also applied in many other and very different fields, e.g. theoretical biology \cite{KellerChristen2018}, medicine \cite{Ohs2016,Christen2018}, and quantum mechanics \cite{Hooft2016}. Agent-based models, on the other hand, are currently mostly used in the social sciences studying the behaviour of agents such as pedestrians moving in a subway station \cite{Chen2017}. With the availability of large training data sets and computing power, artificial neural networks and other artificial intelligence methods are widely used today in many applications, from the classification of skin cancer \cite{Esteva2017} to the solution of the quantum many-body problem \cite{Carleo2017}. Although current artificial intelligence methods seem to be general enough to apply them to many applications, they are still limited to certain problems and impose requirements on the available data and the interpretability of the solutions. Evolutionary algorithms are another type of model that has been successfully used for problems where the data might be incomplete, or an optimisation with multiple objectives is required \cite{Bezerra2016}. They have been also suitable for searching for model parameters in combination with other computer models \cite{Bidlo2016}.

This indicates that there are specific computer models for certain problems and applications rather than a metamodel that is able to tackle any problem and in any application without limitations. It also means that by choosing a particular computer model, we might limit the possible solutions and their interpretability. E.g. if a cellular automaton is used, the solution will be represented by the way the cellular states are updated over time with a transition function. If an artificial neural network is used, the solution will be represented by the weights that are generally difficult if not impossible to interpret. This is referred to as the interpretability problem of current methods of artificial intelligence and is especially critical in applications such as medicine. It seems, however, that these methods have a large solution space and thus can potentially come up with creative solutions. Models providing more interpretable solutions, in contrast, seem to have a smaller solution space limited by the model itself. An alternative way to choose models or to build them is therefore proposed in the present study, which seeks to overcome these restrictions on possible solutions and their interpretability.

We argue that such an alternative option could be the automatic programming of computer models based on a metamodel where as little limitations as possible are imposed on the model creation. The input data would guide the concretisation of a suitable model from the metamodel. Possible solutions or models could be programmed from certain code blocks that are meaningful to humans and combined together in an evolutionary computation. In extreme cases, i.e. without limitations, program code would be written automatically by choosing symbols that are valid in the respective programming language and then models would evolve based on choosing and putting together these symbols to create program code. One can imagine that this would generate program code that would almost always fail when compiling. In addition, even if it were to compile or the compilation problem could be reduced with programming languages such as LISP or interpretative languages, it is still hard to imagine that this would create readable program code and thus interpretable for humans. Instead of allowing any valid symbol, one might limit the choice to certain model building blocks that can be chosen, adapted, and combined in an evolutionary computation. If, in addition, these model building blocks are defined in such a way that they make sense for humans, the evolutionary computation would potentially be able to develop novel solutions or models that can be interpreted by humans and where only the input data imposes restrictions.

However, this is a large problem in which, in a first step, the questions need to be addressed how to define model building blocks for automatic programming of computer models, and how to implement and combine them in a running computer program. The first question has been recently addressed in detail in another study by the authors \cite{ChristenDelFabbro2019}. The purpose of the present study is to answer the second question based on philosophical concepts and model building blocks from the previous study. Philosophers have thought for centuries about the structure and behaviour of objects and phenomena of many kinds. We proposed in the earlier study to proceed in the same way, particularly by using concepts developed by the French philosopher Gilbert Simondon. Via paradigms and schemes borrowed from cybernetics \cite{Wiener1948,Ashby1956}, his philosophy describes the functioning of what could be called complex systems today, especially technical objects such as engines but also natural processes such as crystallisations \cite{Simondon2013,Simondon2016}. Therefore, his method seems closely related to computer modelling of complex systems that wants to achieve the same. With a slight deviation in the use of these concepts, we have integrated Simondon's concepts directly into the functioning of a metamodel for complex systems, where it uses philosophical concepts as a guideline for computation. In this sense, one could speak of philosophical computation. The aims of this study are therefore to propose an approach on how the previously defined model building blocks could be implemented and combined into computer models, and to provide first experiments of automatically programmed cellular automata and artificial neural networks from the metamodel.

\section{The Metamodel}

In this section, we describe the metamodel for complex systems and its model building blocks as developed in the earlier study \cite{ChristenDelFabbro2019}. These building blocks were developed according to the philosophical concepts structure, operation, and system as defined and described by Simondon \cite{Simondon2013}.

\subsection{Structure}
Structure represents the spatial or topological dimension of the system, e.g. agents, cells, nodes, lattices, or grids. In this configuration, no processes and no dynamics are active. Structure represents the topological configuration of the system in its most basic  arrangement.

\subsection{Operation}
Operation represents the system's temporal dimension. It considers the behaviour of the system, e.g. it looks at how cells, agents, or nodes control and affect each other. Moreover, on a more fundamental level, it also describes how structures dilate topologically over time, e.g. how single cells, agents, or nodes are formed initially. Similar to drawing a straight line on a sheet of paper, the spatial dimension of the line is formed at the same time as the temporal operation of drawing moves on \cite{Simondon2013}. In summary, operation defines how agents are generated temporally and, once formed, how they behave in their specific environment.

\subsection{System}
However, in reality, no system is composed by operations or structures alone, i.e. every system has a spatial \textit{and} a temporal dimension. Hence, a \textit{system} is defined by the product of structure and operation, without any concrete parametrisation.
Once parametrisation begins, structure and operation find themselves in a so-called \textit{metastable system}. This means that in metastability more and more parameters such as initial conditions of the parameter states and the dynamics of the modelled system are defined in order to have a computable model. In this sense, metastability represents a transitional regime, which is at the same time partly virtual and actual. Hence, structure and operation are initially defined in a virtual regime and while more and more parameters are included, such as initial conditions and dynamic update functions, the model itself becomes more and more concrete in metastability and finally computes in actuality (Fig.~\ref{fig:Figure1}).

\subsection{Model Building Blocks}
These philosophical concepts were built into model building blocks of a generic computer model that are fundamental to every computer model. They are therefore independent from the concrete models such as cellular automata and artificial neural networks and can be regarded as components of the metamodel.

\begin{figure}
  \includegraphics[width=1.0\linewidth]{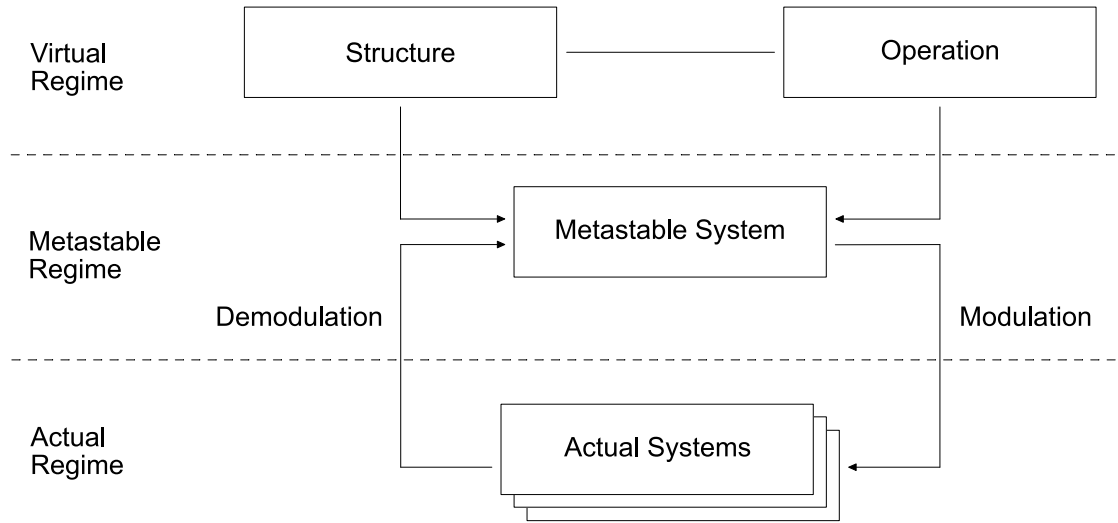}
  \caption{Gilbert Simondon's philosophical concepts applied to meta-modelling of complex systems.}
  \label{fig:Figure1}
\end{figure}

On an abstract level, computer models have at least one structure and at least one operation that are described in a so-called virtual regime. The virtual regime can be regarded as abstract descriptions of the spatial and temporal dimensions of a model (Fig.~\ref{fig:Figure1}).

Formally, structure has been described by a $p$-tuple $e$ of $p$ basic \textit{entities} such as cells, nodes, agents, or elements forming a topology such as a lattice, grid, or network. The environment, neighbourhood, or \textit{milieu} of an entity describing its interaction with neighbouring entities is defined in a $q$-tuple $m_i$ consisting of $q$ neighbouring entities for the $i$-th entity $e_i$ in $e$. The milieus of all entities in $e$ can furthermore be described by the matrix $\mathbf{M}$, which is structured as an adjacency matrix \cite{Cormen2009}.

Operation, on the other hand, can be described with an \textit{update function}
\begin{equation}
\phi:s^{q+1} \rightarrow s,
\end{equation}
where the states of each entity $e_i$ are updated over time. An entity can have a certain state, $e_i \in s$, where $s$ defines the set of possible states. The new state of an entity $e_i^{(t+1)}$ at the next time step $t+1$ is determined based on the states of the entity $e_i^{(t)}$ itself at the current time step $t$ and the states of the neighbouring entities in $m_i^{(t)}$ at the current time step $t$.

The model building blocks entity, milieu, and update function are the basic components of the metamodel. They were described in formal and abstract form. As they represent structure and operation, combining them can form a system, which can be interpreted as a model. If concrete parameters are fed to the system, a metastable system is created, which is then capable of acting in the actual regime. It is the initialisation of the model with given parameters ($\phi$, $\mathbf{M}$, $e^{(t=0)}$, $s$) while the execution of the model afterwards occurs in the actual regime (Fig.~\ref{fig:Figure1}).

In the interjacent regime of metastability two important functions are at work in regard to concrete parameters: modulation and demodulation. On the one hand, modulation superimposes all parameters and prepares thereby the computation of an actual system. On the other hand, demodulation differentiates between structural and operational parameters (Fig.~\ref{fig:Figure1}).

\section{The Automatic Implementation}

The philosophical concepts of structure, operation, and system need to be represented in program code to create automatically generated programs of computer models. The developed model building blocks entity, milieu, and update function describing the metamodel on an abstract level are in the following used to automatically map and implement the philosophical concepts into program code.

\subsection{Representation}
Structure, and therefore the entity tuple $e$ and the milieu matrix $\mathbf{M}$ are well represented with arrays or similar data containers. Generic programming is used to generically define their data type, and dynamic array containers are used to dynamically define their size. In this sense, the metamodel is independent of the type as well as the size of input data provided by the system to be modelled. It allows creating a model of the system that can be concretised in terms of data type and size by feeding concrete parameters into it to form the metastable system. We used the \texttt{vector} template in C++ for the present implementation but other dynamic array containers such as the \texttt{ArrayList} class in Java would also be possible to use.

Operation and therefore the update function $\phi$ describes the temporal dimension defining how the entities $e^{(t)}$ at time point $t$ change their states to the next time point $e^{(t+1)}$ at $t+1$ based on their current states $e^{(t)}$ and the states of the entities in their respective milieu as described in $\mathbf{M}$. Mathematical functions can be represented by functions and methods in procedural and object-oriented programming, respectively. Abstract methods could be used to define these methods in the metamodel, or classes for specific implementations could be used. For simplicity reasons, we implemented specific classes for cellular automaton and artificial neural network models in the present implementation. The parameters are defined in a generic and dynamic way given by the structure implementation. The method body, on the other hand, is specific to the application and therefore is implemented in specific classes that can be used by the metamodel to create concrete systems or models.

\subsection{Implementation}
In the present study, we used an object-oriented programming language to implement the first experiments with cellular automata and artificial neural networks. The reason for this choice was twofold: first, classes allow the abstract and complementary description of structure and operation in the virtual regime and objects allow implementing the metastable regime through initialisation. Second, they often provide dynamic and generic types. C++ was used for the implementation, but any other object-oriented programming language providing these features would have been suitable as well.

The entity tuple $e$ and the milieu matrix $\mathbf{M}$ were implemented with the \texttt{vector} template as data members of a class \texttt{System}. $e$ contains objects of the class \texttt{Entity} with the member data \texttt{state} while $\mathbf{M}$ contains either Booleans to represent relationships only or floating point numbers to represent weighted relationships between entities. The update function $\phi$ was implemented with a method in a specific class. In these first experiments, the classes \texttt{CA} and \texttt{ANN} were created, each implementing a respective \texttt{updateFunction}.

\subsection{Automation}
The present implementation allows to automatically program computer models by specifying the concrete parameters in different ways concretising entities and their milieus as well as choosing $\phi$ from different classes. It is therefore possible to provide all concrete parameters and compute the structural and operational evolution of the system or to provide them only partially varying unknowns in an evolutionary computation. In the present study, as a first step, we varied a part of the update function $\phi$ through an evolutionary computation in the cellular automaton and through a learning rule in the artificial neural network to find a given state of the system.

Automatic programming was achieved through source code generation and writing it to a file followed by programmatically compiling and running the code using \texttt{system()} in C++. The metamodel allows creating different models in the source code generation where a string containing the source code is determined. Variables, model building blocks, and specific update functions are encoded as separate string variables concatenated by the main program. The advantage of a metamodel is evident because the different model building blocks are compatible with each other independently from a specific model implementation and are therefore more easily defined in this way of creating source code. In preliminary tests, we also used the Java Compiler API as well as running it based on the concept of reflection. It led to a rather complicated programming and a code prone to errors. In addition, it is a time-consuming way to solve the current problem. We therefore favoured the less complicated and much faster C++ implementation.

\subsection{Computation}
We have seen that with the transition from the metastable to the actual regime, concrete parameters concretise more and more the actual system. Hereby, further parameters can be categorised, i.e. within structure, the number of entities $p$ and their set of possible states $s$ and the concrete states $e$ define in total the initial conditions $e^{(t=0)}$. On the other hand, within operation, specific dynamic rules, boundary conditions, and the size of the milieu $q$ define in total the update function $\phi$. If all these single parameters are known, then we have a complete system. However, if there are certain unknowns, the system is incomplete. Therefore, one is allowed to conclude that as soon as unknown parameters occur, demodulation processes arise, in which structural and operational parameters can be differentiated. On the contrary, as we have seen above, if concrete parameters are brought together, modulation processes are at work, since structural and operational parameters are more and more superimposed (Fig.~\ref{fig:Figure1}).

To search for these unknown parameters, several steps have now to be taken. First, a problem has to be defined, i.e. one or several unknown parameters have to be defined as such, which then become the subject of investigation. This also means that while several concrete parameters are unknown, several others are known. Therefore, in a second step, new actual systems can be created by modulation starting from these known parameters and by giving the unknown parameters concrete values, e.g. in an evolutionary computation giving the asked parameters random values. It is possible to automatically implement these modulation processes with the present approach. In the main program, the given structures are declared and then used to instantiate a respective \texttt{System} object via the constructor, creating the virtual regime. A first system is then created by initialising known parameters and setting the asked parameter with a random value, which is the transition from the virtual to the metastable regime. The \texttt{updateFunction} method is then repeated until the system leads to acceptable results in the given iteration or until a maximum number of time steps is reached. In a third step, demodulation processes are required to check whether the unknown parameters have been found. Thus, every time a certain number of time steps have been computed, the concrete parameters of the newly obtained system are disaggregated via demodulation. In a fourth step, the new parameters are then compared to the ones that were initially searched for. If there is no correlation or the result is not satisfying, certain values within the initially unknown parameters are changed in order to compute further modulation processes, which then are subjected again to demodulation and comparison. As a result, if one searches for certain unknowns, demodulation and modulation processes are applied interchangeably. The comparison of new and old parameters is required to evaluate whether a system reveals an acceptable result. This is achieved by calculating the match or overlap with the target, quantified as a number between 0 and 1, which is similar to the concepts of fitness in evolutionary computations or loss function in artificial neural networks. The closer it gets to 1, the more significative the new system is and the closer it is to the unknown parameters searched for. The search of these unknowns can be and is in the present work achieved with an evolutionary computation in a cellular automaton experiment and with a perceptron learning rule in the artificial neural network experiment. However, it is important to notice the difference that here the model is created automatically and from abstract concepts, while in the usual case the model is given and implemented and executed manually.

\section{The First Experiments}

In the following, the basic building blocks of the metamodel are concretised in a cellular automaton and an artificial neural network. 

\subsection{Cellular Automaton}
A cellular automaton consists of a discrete cellular state space $\mathcal{L}$ on which the dynamics unfolds, a local value space $\Sigma$ that defines the set of possible states of an individual cell, boundary conditions, and a dynamic update rule $\phi$ defining the dynamics and thus temporal behaviour \cite{Ilachinski2001}. In the present study, we considered a two-state one-dimensional cellular automaton. It is made of cells that are all identical, and periodic boundary conditions were assumed \cite{MainzerChua2012}. In the one-dimensional case, this boundary condition connects both ends of the lattice $\mathcal{L}$ together forming a ring. A Boolean state was furthermore used where the cell $c_i$ at position $i$ in $\mathcal{L}$ can have one of two possible states, $c_i \in \Sigma = \{ 0,1 \}$. Each cell at time point $t$ is updated by a dynamic update function or local transition function
\begin{equation}
c_{i}^{(t+1)}=\phi(c_{i-1}^{(t)},c_{i}^{(t)},c_{i+1}^{(t)}),
\end{equation}
depending on its current state $c_i^{(t)}$ as well as the states of its neighbouring cells $c_{i-1}^{(t)}$ and  $c_{i+1}^{(t)}$. Therefore, there are $n_\Sigma=2$ possible cell states and three input parameters, which leads to $2^3=8$ possible update rules in a two-state cellular automaton with a neighbourhood defined by the nearest neighbours.

Structure is therefore represented by the fact that there is a lattice of cells and that spatially a further cell is situated in the next time generation at $t+1$, e.g. below the considered cells at $t$. Operation, on the other hand, is represented by the fact that within the considered neighbourhood in the lattice, each cell is initially formed and thereby directly linked to its adjacent cell as well as that these cells inform the state of the cell at $t+1$. Only in combination, structure and operation are able to form a system. This system can then be concretised into a metastable and finally actual system or in this case a real computable cellular automaton, by feeding concrete parameters into the system and executing it.

In this first experiment, the dynamic rules were set randomly to automatically generate systems or metastable cellular automata with various update rules. The idea therefore is to generate metastable systems and automatically program them with various update rules until one of them produces a given target output system. This target system was provided, and its specific values were assumed to be the output of Wolfram's rule 110 at $t=15$ \cite{Wolfram2002} while the update rule was searched with an evolutionary computation assuming only one possible solution at a time that is randomly assigned without any selection criterion. With $\phi(c_{i-1}^{(t)},c_{i}^{(t)},c_{i+1}^{(t)})$, rule 110 is the following update function: 
\begin{equation}
\begin{aligned}
\phi(0,0,0) & = 0, &
\phi(0,0,1)	&= 1, &
\phi(0,1,0)	&= 1, &
\phi(1,0,0)	&= 0, \\
\phi(0,1,1)	&= 1, &
\phi(1,0,1)	&= 1, &
\phi(1,1,0)	&= 1, &
\phi(1,1,1)	&= 0.
\end{aligned}
\end{equation}
Starting with the initial configuration of
\begin{equation}
e^{(t=0)}=0000000000000001000000000000000,
\end{equation}
the output or target is
\begin{equation}
e^{\text{target}}=1101011001111101000000000000000. 
\end{equation}
Although the used cellular automaton model is very simple, rule 110 shows exceptional and interesting behaviour that is complex and could be considered on the edge of order and chaos \cite{Kauffman1993}. It is even computationally universal, i.e. capable of running any given program or algorithm \cite{Cook2004}.

In a system where the structural parameters, i.e. the number of entities $p$ and their concrete states $e$ are known, the defined problem is to search for the operational parameters, i.e. a specific dynamic rule.

Structures of type Boolean were implemented representing two states $\Sigma = \{ 0,1 \}$ and 31 entities were modelled. Operation was implemented as a truth table of three input parameters representing the nearest neighbours and the considered cell itself determining one output according to the specified dynamic rules.

The experiment was repeated several times, in each case revealing Wolfram's rule 110 as a solution after less than 1000 attempts of building and automatically programming a metastable and actual system with randomly generated rules.

\subsection{Artificial Neural Network}
This model consists of a network of artificial neurons called perceptrons, where typically input neurons are fed with data and connected with middle or hidden layers of neurons, which in turn are connected to other hidden layers before they are finally connected to the output neurons that provide the result. Different network topologies are therefore possible. Each perceptron $j$ consists of an input function $in_j$ that calculates a weighted sum of its incoming activation signals $a_i$ from perceptrons $i$. The value of the input function is then used in the activation function $g$ to calculate the outgoing activation signal $a_j$ of the perceptron $j$. The output or result of an artificial neural network is thus computed by calculating the activation of each individual perceptron from layer to layer until the output neurons are reached. This type of network is called multilayer feedforward artificial neural network. The weights of the network are then adjusted through a learning rule.

In our experiment, we automatically programmed a multilayer feedforward artificial neural network based on the metamodel with entities, milieus, and an update function. A neuron or perceptron is represented by an object of the class \texttt{Entity} and the milieus by the adjacency matrix $\mathbf{M}$ also storing the weights. The entity tuple $e$ therefore contains all the perceptrons with their current states and the milieu matrix $\mathbf{M}$ the neighbourhood or topology of the network. Comparable to the cellular automaton experiment, we used the same target system $e^{\text{target}}$ and 15 layers, each with 31 perceptrons. The update function consists of the input function $in_j$ and the activation function $g$ calculating the activation signal $a_j=g(in_j)$. In the input function
\begin{equation}
in_j=\sum^n_{i=0}\omega_{i,j}\cdot a_i,
\end{equation}
the $n$ incoming activation signals $a_i$ of neuron $j$ are weighted through a respective weight $\omega_{i,j}$ and summed up. Please note that $a_0$ is a bias weight. The activation function was defined as a threshold function
\begin{equation}
g(in_j)=
\begin{cases}
0 \text{ if } 0.5 > in_j\\
1 \text{ if } in_j \geq 0.5
\end{cases}
\end{equation} 
and the perceptron learning rule
\begin{equation}
\omega_{i,j} \leftarrow \omega_{i,j} + r(y-a_j)a_i
\end{equation} 
was used, where $r$ is the learning rate, $y$ the target or desired activation signal, $a_j$ the current outgoing activation signal, and $a_i$ the current incoming activation signals \cite{RusselNorvig2010}. It is implemented as a method in the \texttt{ANN} class.

Here, too, the experiment was repeated several times, generally requiring less than 100\,000 attempts to build and automatically program a metastable and actual system. The exact solution was not found in any of the tested cases but a match of approximately 90\% was achieved comparing the solution with target entity states.

\section{Discussion}

The aim of the present study was to make use of a previously defined metamodel of complex systems for automatic programming of computer models and to generate and automatically program cellular automata and artificial neural networks. Model building blocks of this metamodel were defined based on the concepts of structure and operation in a previous study \cite{ChristenDelFabbro2019}. It might not be surprising that structure is formally described with mathematical tuples and has thus been implemented with some kind of array or list data structure. Philosophy, however, allows novel and creative guidance. Hence, the philosophical concepts borrowed from Simondon imply a system that consists of some local elements with connections to each other. On this basis, we defined the model building blocks entity and milieu. In line with this, from a temporal perspective, structure can be of different sizes and entity states can be of different types. Therefore, the data structure has to be implemented dynamically and generically. In addition, due to the philosophical definition of operation, it is not surprising to formally describe it with a mathematical function and thus to implement it with a method. Here, the usefulness of philosophy is to allow the operation to change structure. According to the two philosophical concepts of structure and operation, a system is created by superimposing them. It is important to highlight that structure and operation are always interconnected. While structure provides the spatial dimension for the operation to occur, operation, on the other hand, forms the evolving structure and thus defines the connections between the entities. Hence, input and output parameters of the update function have to be implemented and defined, which is part of the model building block operation. This function is application specific, i.e. it represents the functioning of the concrete system. Hence, philosophical concepts such as structure and operation can guide the definition of a metamodel for complex systems. Because of their abstractness, they suggest generic and object-oriented programming for their implementation in the virtual regime.

Moreover, starting from the abstract definition of model building blocks, additional concrete models are formed in a metastable state. Hereby, philosophy provides a framework on how to concretise the abstract model building blocks into a concrete computer model. Concrete parameters are fed into the metastable system. These concrete parameters are the initial conditions and an update function, which in turn inform the metastable system. This guides the implementation by using the concrete parameters to initialise a \texttt{System} object, which then represents the metastable system.

Since structure and operation are complementarily interrelated, it is important to emphasise that in the virtual regime, structure and operation have neither a categorical nor a hierarchical relationship. Therefore, every structure is operated, and every operation is structured. While concrete parameters are being fed into the system, thereby transitioning from the virtual to the metastable regime, the virtual regime itself is not being altered. Hence, it is the starting point in order to create constantly new metastable models within the metastable regime. By accepting the virtual regime as underlying governor, computation itself is constantly shifting from the metastable to the actual regime producing new systems, new models. This also means that modelling itself happens only within the field of concrete parameters. It is not re-entering the virtual regime of abstractness. Spoken philosophically: The starting point as much as the result of every computation is always a model or some kind of image. Even if the majority of concrete parameters is unknown, it should be possible to compute different possible outcomes, simply by giving the unknowns random values, e.g. in an evolutionary computation.

Because of the complementary relation of structure and operation and its unfolding in metastability, computation of actual systems is not reduced to the meta-level of the virtual regime. Our purpose was to create a \textit{method}, which is able to create models of complex systems.  In this sense, our method can be seen as a tool rather than a theory. We call this method the allagmatic method, because of its relatedness to Simondon's work \cite{Simondon2013}. For Simondon, structure and operation are generic concepts, which are able to describe all sorts of systems in reality. Allagmatic is derived from the Greek verb allatein, meaning change, transition, or transformation. Hence, structure and operation are not only complementary, they also influence each other reciprocally. Based on this relationship between abstract and concrete computation, the allagmatic method is highly adaptable. Since the virtual regime, except from its spatial and temporal constitution is not fixed to any specific type of model or image, it can potentially undermine all types of computer models. As we have shown, the metamodel allows to produce different types of models, such as cellular automata and artificial neural networks, if not novel kinds of models that are still undiscovered. 

However, with the immense adaptive behaviour comes also a high degree of freedom, which has to be controlled in order to use the allagmatic method in a pragmatic way. This control is achieved by defining initial values based on the problem to be solved. Yet, since computation always happens within the metastable and actual regime, three problems can occur: a) having a totally incomplete system, i.e. all parameters are unknown, b) having a semi-complete system, i.e. certain parameters are unknown and certain are known, and c) having a totally complete system, i.e. all parameters are fully known. Even if at first sight a problem of kind a) seems meaningless, it might be fruitful if one considers the possibility of the allagmatic method to create totally new types of models. Here, it would be possible to explore the possible unfolding of the computational universe \cite{Wolfram2002}. A problem of type b) was the initial scenario of the presented experiments; not only could it be widened by increasing the number of unknowns, it could also be modelled by a different kind of model such as an agent-based model. Problems of type c), where all parameters are fully known, can be looked at as being subjected to possible predictions. Starting from fully known parameters, future possible outcomes can be computed and evaluated. Notice that it is also possible for problems to transmute, i.e. if starting from a problem b) by knowing only certain parameters and with the help of computation the unknown parameters are found, one ends up with the situation of problem c) where all parameters are known and where from now on predictions of the behaviour of the system can be made.

We also provided first examples of formal descriptions as well as automatic implementations of a two-state one-dimensional cellular automaton and a multilayer feedforward artificial neural network using the allagmatic method. What is usually defined as a cellular automaton cell is built from the abstract entity model building block. These cells form a lattice, which is determined by the entity tuple $e$. Typically, this lattice would be implemented with an array and the milieu would not be stored for each cell as this is always defined by the neighbours and thus given from the grid structure in a cellular automaton. Here it becomes evident that the traditional way of implementing a specific model is more efficient but at the same time it is also more restrictive with respect to the topology of connections between entities. Our approach does not have this limitation but comes with a less efficient implementation. This is not an issue in the present study since the aim is to find a way to build and automatically program computer models with as few limitations as possible.

In addition to cellular automata and artificial neural networks, the metamodel defined in the allagmatic method arguably also allows to build other computer models such as agent-based models. Agent-based models are closely related to cellular automata and as such have a similar structure and operation. Even more so, they have agents and agent behaviours, which are well represented by the model building blocks entity and update function, respectively. Agents also communicate with each other from which the behaviour of the whole system emerges. This is well represented by a model building block, the milieu and output of the update function. Maybe even more importantly, the generality of the model building blocks possibly allows creating novel computer models still not known to us. The created models therefore are no longer necessarily of one known model type, they might only show certain features of them. They blur the line between distinct computer models and thus are able to produce novelty. In addition, these newly generated computer models can possibly also be interpreted by humans. They consist of the model building blocks that can be related to the philosophical concepts, which provide a description of the role of each model building block and how they are related to each other.

\section{Conclusion and Outlook}

We conclude that automatic programming of computer models can be achieved by a previously developed metamodel \cite{ChristenDelFabbro2019} that was guided by philosophical concepts such as structure and operation, particularly on an abstract level. Other concepts such as metastable system and concrete parameters are also useful for building concrete computer models and for executing them. Both the abstract and the concrete definitions provide guidance for mathematical description and automatic programming of computer models.

While we provided exemplified descriptions and automatic programming experiments of simple cellular automaton and artificial neural network models, further studies are required to formalise and implement other models, especially agent-based models and artificial intelligence methods. Because of the generality of the building blocks in the metamodel of the allagmatic method, it is likely that the formalisation and automatic programming of models other than cellular automata and artificial neural networks are possible. This will allow to not only automatically program existing and novel computer models, it will also create models that are interpretable and thus understandable to humans. Such novel but still interpretable computer models will finally help us to more deeply explore and understand the computational universe.

\section*{Acknowledgements}

This work was supported by the Hasler Foundation under Grant 18067.

\bibliographystyle{unsrt}
\bibliography{References_Automatic_Programming.bib}

\end{document}